\documentclass{article}

\usepackage{arxiv}

\usepackage[utf8]{inputenc} 
\usepackage[T1]{fontenc}    
\usepackage{hyperref}       
\usepackage{url}            
\usepackage{booktabs}       
\usepackage{amsfonts}       
\usepackage{nicefrac}       
\usepackage{microtype}      
\usepackage{lipsum}
\usepackage{graphicx}
\graphicspath{ {./images/} }

\usepackage{natbib}
\usepackage{listings}

\title{Creating an Aligned Corpus of Sound and Text: \\ The Multimodal Corpus of Shakespeare and Milton}

\author{
Manex Agirrezabal \\
Centre for Language Technology (CST)\\
Department of Nordic Studies and Linguistics (NorS)\\
University of Copenhagen\\
2300 Copenhagen\\
Denmark\\
  \texttt{manex.aguirrezabal@hum.ku.dk} \\
}

\begin{document}
\maketitle
\begin{abstract}
In this work we present a corpus of poems by William Shakespeare and John Milton that have been enriched with readings from the public domain.  We have aligned all the lines with their respective audio segments, at the line, word, syllable and phone level, and we have included their scansion. We make a basic visualization platform for these poems and we conclude by conjecturing possible future directions.
\end{abstract}

\keywords{poetry \and multimodal \and text \and audio \and scansion \and phones \and syllables} 

\section{Introduction}
\label{intro}

Poetry is part of any culture around the world. Poems have been used to transmit knowledge from generation to generation.\footnote{See for example the Aryabhatiya poem at \url{https://en.wikipedia.org/wiki/Aryabhatiya}.} These poems can take the shape of a nursery rhyme, a traditional proverb, and so on.
Currently, if we want to analyze poems, there are a number of corpora concerning poetry, such as The Gutenberg English Poetry corpus \citep{jacobs2018gutenberg}, or the similarly called Gutenberg Poetry Corpus,\footnote{\url{https://github.com/aparrish/gutenberg-poetry-corpus}} compiled by Allison Parrish. Our goal is, though, to understand the underlying relation between text and audio, and then, for instance, we could ask how is a specific sound or stress pattern acoustically realized. One can wonder about the relation between rhymes and the pitch of our speech. Our main goal is to make it possible to investigate this relationship by providing relevant resources to the research community. To this end, we have collected and aligned the work and respective readings of two classic English poets, William Shakespeare and John Milton.

If we think about poetry, rhythm and sound, we will find ourselves diving into what is called \textit{scansion}. The task of scansion involves the analysis and marking of rhythmic structures in poems, together with the analysis of rhymes.
Traditional scansion also includes dividing lines into sequences of repeating patterns, namely feet. Consider, for instance, a portion of the poem \textit{The Passionate Shepherd to His Love} by Chris Marlowe:

\begin{center}
\textit{Come live with me and be my love,}\\
\textit{And we will all the pleasures prove,}\\
\textit{That Valleys, groves, hills, and fields,}\\
\textit{Woods, or steepy mountain yields.}\\
\end{center}

\noindent In this poem, we can observe that there is a strong beatlike rhythm when we read the poem aloud (``\textit{Come \textbf{live} with \textbf{me} and \textbf{be} my \textbf{love}}'').

Let us consider another example. Please check the lines in Figure \ref{scansion:milton}, which are independent lines from the 9th book of Paradise Lost by John Milton \citep{pickering_poetical_1832}. The rightmost section in each line shows the prominence that should be made in those lines according to an automatic scansion model. The automatic scansion model is a publicly available scansion model that given raw text, it returns the stresses that should be realized. Even though the model makes some mistakes, we can say that it captured the meter of the poem fairly well. Furthermore, we would like the reader to check the highlighted word ``with''. The scansion model has captured that beyond its pure positioning in the line, the prominence of that word may vary, based on the context it appears.  In the second line, the first appearance of the word is unstressed and the second one is stressed (\emph{with \textbf{man} as \textbf{with} his \textbf{friend} fa\textbf{mi}liar \textbf{used}}).

Several researchers have worked on scansion, where, given the text of a poem, stresses were marked. Scansion can guide us with information on how a poem is expected to be read, but are these expectations realized? And how are they manifested? We endeavor to address queries of this nature.

\begin{figure*}[ht]
\begin{center}
\begin{tabular}{l|c|l}
 (1) & No more of talk where God or Angel guest & \texttt{x/x/x/x/x/}\\
 (2) & \underline{With} Man, as \underline{with} his friend, familiar us'd, & \texttt{x/x/x/xx/x/}\\
(3) & To sit indulgent, and \underline{with} him partake & \texttt{x/x/xx/x/x}\\
\multicolumn{3}{c}{\ldots}\\
(73) & Rose up a fountain by the tree of life: & \texttt{/xx/x/x/x/}\\
(74) & In \underline{with} the river sunk, and \underline{with} it rose & \texttt{/xx/x/x/x/}\\
(75) & Satan, involved in rising mist; then sought & \texttt{/x/x/x/x/x/}\\
\end{tabular}
\end{center}
\caption{Excerpts from the 9th book of \textit{Paradise Lost} by John Milton and its scansion (automatic)}
\label{scansion:milton}
\end{figure*}


In this work we present a multimodal dataset\footnote{\url{https://github.com/manexagirrezabal/shakespeare_milton_multimodal}} that can help us in answering the above questions, but can also lead to new intriguing research questions. We have created a corpus with poetry by Shakespeare and Milton, where we save the text and audio, and the text is expressed as raw text, lines, words, syllables and phonemes. All this information is aligned and for each line we include the output of an automatic scansion model.

The article is structured as follows. First we introduce some related work, considering both similar poetry corpora and also works related to poetry scansion. After that we present the corpus, by mentioning the source of the data, the tools we used and also the encoding of the data. We present and discuss some descriptive statistics about our corpus of study in the next section. Then, we briefly show a website that we made to show the corpus, with some of its characteristics. Last but not least, we conclude the work by proposing some questions that can now be answered and suggest possible future directions.

%
%
    %
    %
    %
    %
    %
    %

\section{Related work}
The two most related projects to the one that we are describing here are \citep{meyer-sickendiek_rhythmicalizer_2017} and \citep{delmonte_sparsar_2013,delmonte_exploring_2016}.

In the project called \textit{Rhythmicalizer} \citep{meyer-sickendiek_rhythmicalizer_2017}, the goal of the authors was to identify rhythmic patterns in free verse poetry. They employed both textual information together with sound, and their final goal was to apply empirical methods for the analysis of such patterns. In later works \citep{meyer-sickendiek_analysis_2018}, the authors perform both feature-based classification and neural classification to predict several common rhythmical patterns present in modern poetry. It should be noted that the sound files that the authors used are recordings made by the poets themselves. Our goal in this paper is to present a corpus similar to the one that they use, but much smaller in number of authors (only two authors) and with less variation in rhythm (more classic rhythms, mostly iambic lines).

In \cite{delmonte_exploring_2016} the authors analyzed Shakespeare's Sonnets using a system called \textit{SPARSAR}, where poems are analyzed regarding different aspects, such as, rhythm, rhyme, meaning, emotion and so on. The authors made a very deep analysis of Shakespeare's sonnets and applied it to produce expressive readings of poems by a Text-To-Speech synthesizer. We think that the corpus that we present in the current work could be complementary to the extensive work by \cite{delmonte_exploring_2016}.






\subsection*{Similar poetry corpora}
There are several poetry corpora available for the computational linguistics community. Relevant cases include \textit{For Better For Verse} \citep{tucker_poetic_2011}\footnote{\url{https://github.com/waynegraham/for_better_for_verse}} in English, the corpus of \textit{Golden-Age Spanish sonnets} \citep{navarro-colorado_metrical_2016}\footnote{\url{https://github.com/bncolorado/CorpusSonetosSigloDeOro}} and \textit{ReMetCa}, the repertory  on Medieval Spanish poetry, \citep{gonzalez-blanco_garcia_remetca:_2013}\footnote{\url{http://proyectoremetca.weebly.com/}} in Spanish and the \textit{Corpus of Czech Verse} \citep{plechac_corpus_2015}\footnote{\url{http://www.versologie.cz/en/kcv.html}} in Czech. Some of these works have been realized as part of the Postdata project \citep{malta_postdata:_2016}.





\begin{figure*}
\scriptsize
\begin{lstlisting}
<TEI version="5.0">
<teiHeader type="text">...</teiHeader>
<media mimeType="audio/wav" url="../audio/shakespeare/shakespeare_sonnet_002.mp3" dur="46.90s">
<desc/>
</media>
<text id="POEM_MARKUP">
 <body>
  <lg n="0">
   <l n="0" beginning="00:00:00.0" end="00:00:03.80" scansion="-+-+-+-+-+">
When forty winters shall besiege thy brow,
    <phones>...</phones>
    <sylls>...</sylls>
    <words>...</words>
   </l>
   <l n="1" beginning="00:00:03.80" end="00:00:07.08" scansion="-+++--+-+">
And dig deep trenches in thy beauty's field,
    <phones>...</phones>
    <sylls>...</sylls>
    <words>...</words>
   </l>
   <l n="2" beginning="00:00:07.08" end="00:00:10.16" scansion="-+++---+-+">
Thy youth's proud livery so gazed on now,
    <phones>...</phones>
    <sylls>...</sylls>
    <words>...</words>
   </l>
   <l n="3" beginning="00:00:10.16" end="00:00:14.12" scansion="-+-+-+-+++">
Will be a tatter'd weed of small worth held:
    <phones>...</phones>
    <sylls>...</sylls>
    <words>...</words>
   </l>
    ...
  </lg>
 </body>
</text>
\end{lstlisting}
\caption{Excerpt of the dataset, from the second Sonnet by William Shakespeare showing only the line level alignments.}
\label{listing:xmlcontent}
\end{figure*}
\normalsize

\section{Shakespeare and Milton: The aligned corpus}
As we are interested in the relationship between the written text and the corresponding uttered sound, we present a corpus of poems where on the one hand we can see poem lines, and on the other hand we can find that same text excerpt read aloud. The whole corpus includes a rough estimate of 12.5 hours of readings. There are almost 100,000 tokens and close to 17,000 types.

This information has been enriched with the text divided into lines, words, syllables and phones, all of which are aligned with the audio information. Literary scholars have analyzed rhythmic patterns in poetry using scansion, and therefore, we decided to perform automatic scansion on all these poems, and to include these analyses. We have used a neural network based approach, as it has been proved to perform the best for this task.

\begin{figure*}
    \centering
    \includegraphics[width=\textwidth]{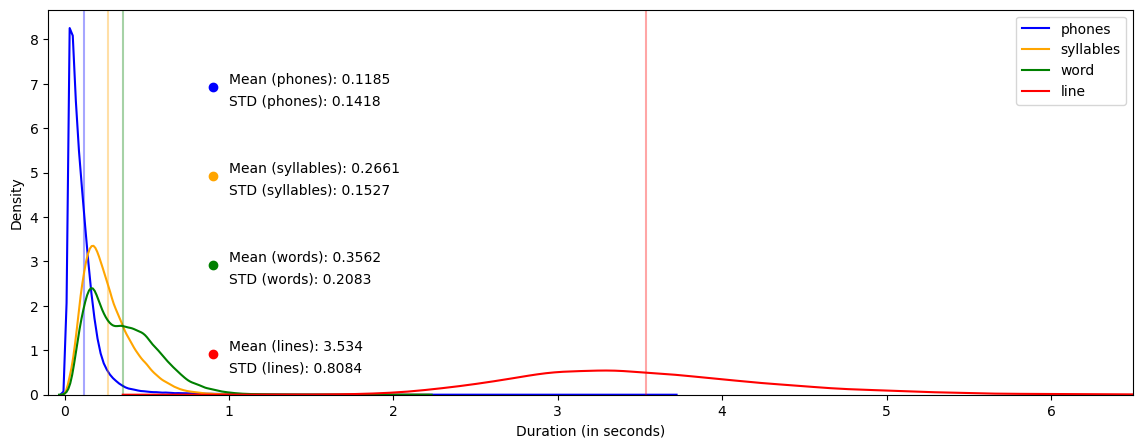}
    \caption{Caption}
    \label{fig:kdeplot_lengths}
\end{figure*}

\subsection{Source of the data}
We use Milton and Shakespeare as the starting point of this work, because they are two extensively analyzed authors. There are available recordings at Librivox and also the textual edition of their work is easily available. Furthermore, this could potentially be used to further extend the work made by \cite{hayes_maxent_2012}.

We obtained the texts from Project Gutenberg \citep{hart_project_1971}. We have used the following document\footnote{\url{http://www.gutenberg.org/ebooks/1041}} to get the text from Shakespeare's Sonnets, and we used this one\footnote{\url{https://www.gutenberg.org/ebooks/26}} for Milton's Paradise Lost. We have used those editions because those were the ones that best matched the audio files.

The audio files were obtained from the Librivox project \citep{kearns_librivox_2014}. We used the reading by Chris Hughes, available in the following link as the audio source for Shakespeare's Sonnets.\footnote{\url{https://librivox.org/shakespeares-sonnets/}} In the case of Paradise Lost, we got the audio files from the reading by Thomas A. Copeland.\footnote{\scriptsize\url{https://librivox.org/paradise-lost-by-john-milton-2/}} When choosing audio version, the only condition that we posed was that each collection of poems should be read by one single person, and not by several people.

\subsection{Processing pipeline}
We initially performed some manual cleaning to remove the initial disclaimer, to make sure to keep only the relevant information for the project. After that, several tools have been used for the construction of this corpus.

Our next step was to have an alignment between sound and text.
The sound files were rather long, especially the recordings of Paradise Lost, where the recordings were, on average, longer than 45 minutes. In such long recordings, it is recommended to perform line-level alignment first, and after that, perform further processing, if necessary. For the sake of consistency, we followed the same procedure for all the recordings. We use Dynamic Time Warping \citep{muller2007dynamic} for aligning each line of the poems to its respective audio section. This algorithm is implemented in  \textit{Aeneas}\footnote{\url{https://github.com/readbeyond/aeneas}} \citep{pettarin_aeneas_2017}, a software that can be used to automatically synchronize text and audio.

We further process the files to perform forced alignment at the phoneme level, for which we have used HTK, an HMM-based based implementation. We used HTK from FAVE \citep{rosenfelder2011fave}, a software to perform forced alignment, which is built upon the Penn Phonetics Lab Forced Aligner. Information about the acoustic models that this software uses can be found in \cite{yuan2008speaker}.\footnote{For the readings of Milton and Shakespeare, a British English forced aligner may have been expected to work better, but it is shown that forced aligners trained on American data are very reliable \citep{MacKenzieTurton2020}.} This alignment model looks up in the CMU pronouncing dictionary \citep{weide_cmu_1998} to guess how the phonemic representation of a word should be. In the case that a word does not appear in the dictionary, we resort to an automatic Grapheme-to-Phoneme tool \citep{g2pE2019}.\footnote{\url{https://github.com/Kyubyong/g2p}}

As important rhythmic elements, we were interested in getting the timing of syllables. To this end, we perform syllabification on phonemic representations of words, and extend the phone-level alignments to the syllable and word level. The syllabification model follows a rule-based method and was adapted from \citet{hulden_finite-state_2006} using the software Foma \citep{hulden_foma:_2009}.

Furthermore, we used a pretrained a BiLSTM+CRF based model\footnote{This model was shown to have an accuracy of 90.80\% at the syllable level, and 53.29\% at the line level. We will provide more specific information when the anonymity period ends.} to perform scansion. This model has been built on top of the work by \cite{lample_neural_2016}\footnote{\url{https://github.com/glample/tagger}} and trained on \cite{tucker_poetic_2011} available on Github.\footnote{\url{https://github.com/waynegraham/for_better_for_verse}}

Then, the pipeline to process the text and audio can be summarized in these 5 steps:

\begin{itemize}
    \item Line-level text and audio alignment using Dynamic Time Warping (DTW)
    \item Grapheme-to-Phoneme conversion using the CMU dictionary or a G2P model
    \item Syllabification based on phone representation
    \item Word/Syllable/Phone level alignment
    \item Automatic scansion of text
\end{itemize}

\subsection{Encoding of data}
We decided to encode the poems following the Text Encoding Initiative \citep{tei_consortium_tei_2008} in its fifth version (TEI 5.0), particularly following the guidelines from the sections about \textit{Verse} and \textit{Transcriptions of Speech} (section 6 and 8, respectively).

In order to encode the alignment of text, its sound, its scansion and the alignments themselves, we first encoded each poem as a line group (\texttt{<lg>}) composed by lines (\texttt{<l>}), as it is commonly done. For each line we include the following information as XML attributes: \texttt{beginning} (moment in which the utterance starts in the audio file), \texttt{end} (the moment in which the utterance ends), \texttt{n} (line number) and \texttt{real} (scansion of the line). Each line is then further divided into phones, syllables and words and for each of these units we include \texttt{beginning} and \texttt{end} as attributes. Please check Figure \ref{listing:xmlcontent} for a general example excerpt of the dataset, and Figure \ref{listing:xmlcontent_deepalignments} in the appendix to understand how aligned data (at the word, syllable and phone level) is stored in the same portion of data.

\section{Some descriptive statistics}
In the following lines we include some descriptive statistics obtained from the analysis of the corpus and the processed data. Based on the beginning and ending attributes from words, syllables and phones, we calculated their average duration and standard deviation, which can be seen in the table below.

\begin{center}
\begin{tabular}{lcccc}
\hline
 & Mean & STD \\
\hline
Phone duration & $0.1185$ & $0.1418$ \\
Syllable duration & $0.2661$ & $0.1527$ \\
Word duration & $0.3562$ & $0.2083$ \\
\hline
\end{tabular}
\end{center}

Each word was divided into syllables using a syllabifyer, but each line was also divided into syllables by the scansion system. The first one is the rule-based syllabification algorithm based on \citet{hulden_finite-state_2006} and the other one is based on an End-to-End scansion model, which given a line of poetry, it returns the rhythm of each syllable as a binary output.
A portion (2\% of the data) of the corpus was manually tagged with the number of syllables and we used that to assess how good both mechanisms were for counting syllables.  We assess both systems in both Shakespeare and Milton and the results can be found in Table \ref{tab:no_syllables_performance}.

\begin{table}[h]
\centering
\begin{tabular}{|l|l|}
\hline
\multicolumn{1}{|c|}{\textbf{Model Configuration}} & \multicolumn{1}{c|}{\textbf{F1-Score}} \\
\hline
RB Syll. (Milton) & $0.85$ \\
E2E Scansion (Milton) & $0.75$ \\
\hline
RB Syll. (Shakespeare) & $0.90$ \\
E2E Scansion (Shakespeare) & $0.80$ \\
\hline
\end{tabular}
\caption{Performance Metrics for Different Models}
\label{tab:no_syllables_performance}
\end{table}

\begin{figure*}[t]
    \centering
    \includegraphics[width=\textwidth]{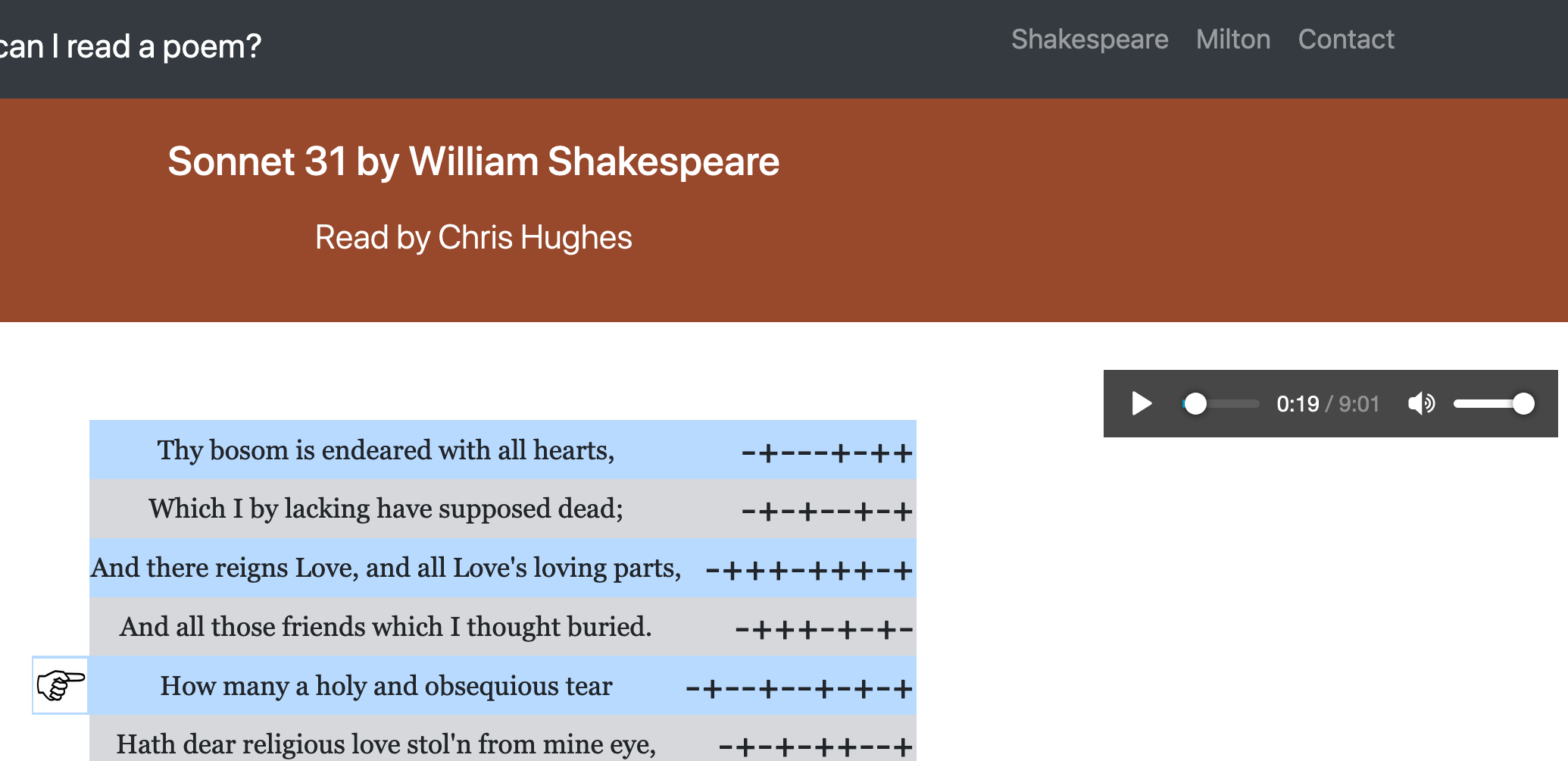}
    \caption{Screenshot of the website, where the finger on the left of the poem shows which line of the poem we are currently listening}
    \label{fig:website}
\end{figure*}{}

Apart from typical evaluation metrics, we investigated how big the errors were for the different models. It seems as the rule-based syllabification model is at maximum one syllable away from the actual number of syllables. The scansion model seems to make bigger mistakes, probably because of the effect of out of vocabulary words. When a word is not recognized, its scansion is automatically returned as a single unstressed syllable, even for very long words. This affects substantially to the number of syllables for long unknown words.

\begin{figure}
    \centering\includegraphics[width=\columnwidth]{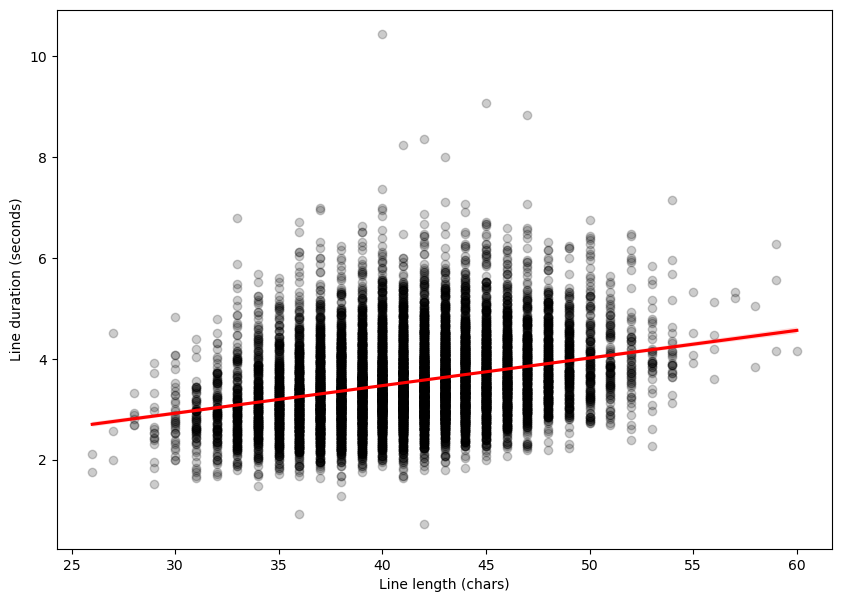}
    \caption{The correlation between line duration in seconds and line length in characters. This shows a Pearson correlation of $0.2932$.}
    \label{fig:linedurlinecharscorr}
\end{figure}

\begin{figure}
    \centering\includegraphics[width=\columnwidth]{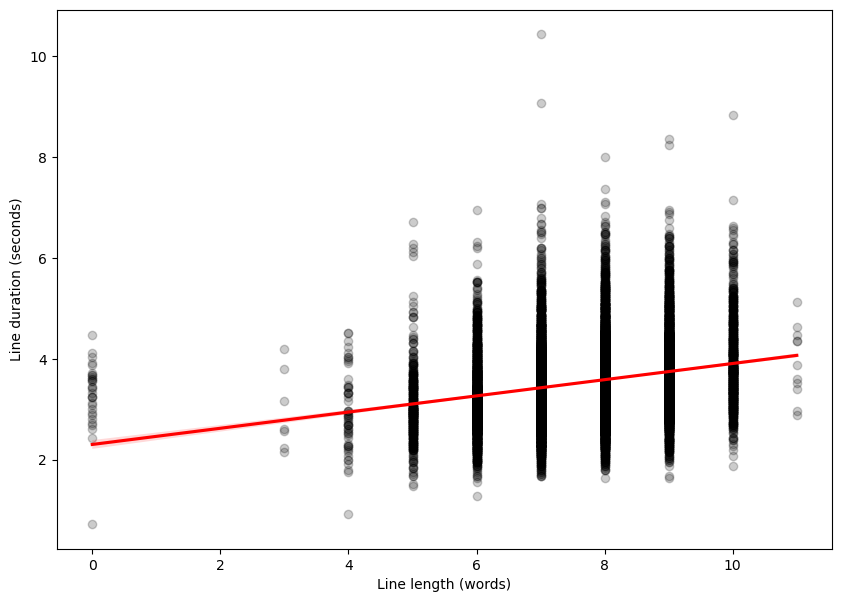}
    \caption{The correlation between line duration in seconds and line length in words. This shows a Pearson correlation of $0.2514$.}
    \label{fig:linedurlinewordscorr}
\end{figure}

\begin{figure}
    \centering\includegraphics[width=\columnwidth]{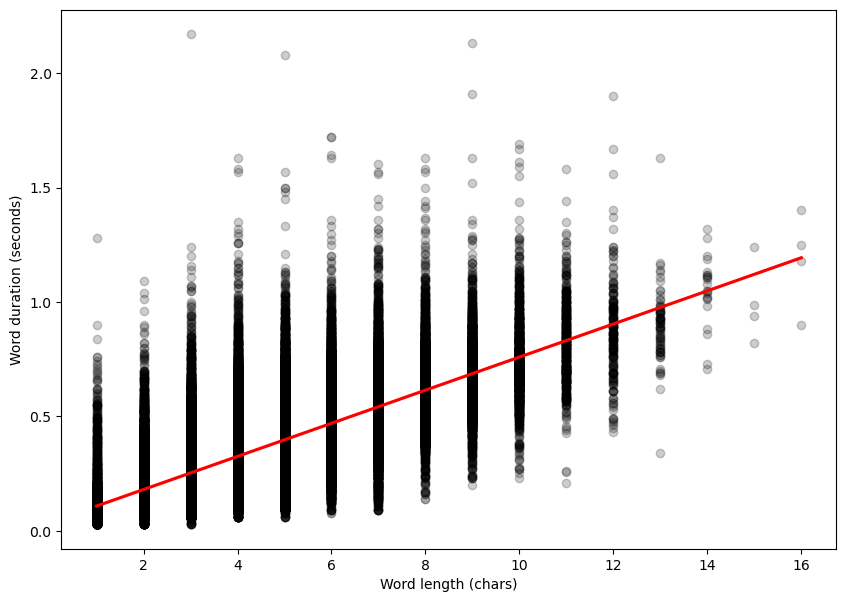}
    \caption{The correlation between word duration in seconds and word length in characters. This shows a Pearson correlation of $0.7163$.}
    \label{fig:worddurwordlengthcorr}
\end{figure}

We also explored what is the correlation between the length of lines and words and their duration. Our main expectation was to observe a positive correlation in all the cases. Figure \ref{fig:linedurlinecharscorr} shows a scatter plot of lines, where we plot the line length in characters in the horizontal axis and the line duration in seconds in the vertical one. In Figure \ref{fig:linedurlinewordscorr} we can see the same information in the vertical axis, but the horizontal one shows the length of the line in number of words. Finally, we can see in Figure \ref{fig:worddurwordlengthcorr} a plot of the word length in characters and the word duration in seconds, in the horizontal and vertical axes, respectively. These last two variables are the ones that show a relatively high correlation, if we compare it to the other two figures, as it can be seen in Table \ref{tab:correlationtable}.

\begin{table}[h]
\centering
\begin{tabular}{|l|c|}
\hline
\textbf{Correlation} & \textbf{Pearson Correlation} \\
\hline
Line (secs. - chars.) & $0.2932$ \\
Line (secs - words) & $0.2514$ \\
Word (secs. - chars.) & $0.7163$ \\
\hline
\end{tabular}
\caption{Correlation between the duration and their lengths in either lines of words}
\label{tab:correlationtable}
\end{table}

It seems rather intuitive that the longer the word (in characters), the longer it will take to be uttered. But this does not happen at the line level, as one may observe in the correlation between line length in characters and line duration in seconds (first row in Table \ref{tab:correlationtable}).

\section{Visualization of the corpus}
We have created a website to show this corpus, as can be seen in Figure \ref{fig:website}. The first step in the application is to choose an author, and after that we have to choose a poem (sonnet number or book number). Once we choose those by clicking, we will get a table that looks like the one in Figure \ref{fig:website}, with two columns. The first column contains a poem line, and the second one has the scansion of that line. As it can be seen, there is also a floating audio player on the right, which will follow going down as we scroll down the poem. When we click (left click) in the text of a poem, we will be able to listen to the poem starting from that specific position. Word, syllable or phone level information is still not visualized, but it should be included in the near future.

\section{Conclusion and Future Work}
We have presented a corpus\footnote{\url{https://github.com/manexagirrezabal/shakespeare_milton_multimodal}} that includes written poems to which we have added audio files from the public domain, and also rhythmical analyses or scansions, by means of an automatic tool. Each line is then enriched with words, syllables and phones, all of them aligned using automatic tools. This multimodal corpus creates new possibilities in the rhythmic analysis of poetry. It could be seen as a corpus that will help reduce the gap between linguistics, literature, acoustics and computer science.

This corpus is going to be used to analyze how poetic stress is realized when reading a poem. But apart from that, the relation between strictly text related asoects and acoustic information could be explored. For instance, can we find evidence of using a higher amplitude (louder voice) when mentioning special characters? When we use words with certain sentiment or polarity, can we observe a change of voice?

If we decide to extend this corpus to include poets that use other kinds of feet, we could check empirically whether the length of each foot, no matter the kind, is the same or not. Consider, for example, the following excerpt from a poem by Dr. Seuss \citep{seuss_scrambled_1953}:

\footnotesize
\begin{center}
\textit{And that's because ever since goodness knows when,}

\textit{They've always been made from the eggs of a hen.}

\textit{Just a plain common hen! What a dumb thing to use}

\textit{With all of the other fine eggs you could choose!}
\end{center}
\normalsize

In this poem, the author uses iambic feet at the beginning of each line, but afterwards they use anapestic feet. If we get a read aloud version of the poem, and we can align it with the text, we could check whether all feet have a similar duration, no matter the number of syllables of a foot. If we were to do that, though, each foot would have to be marked.

\subsection*{Further development of the corpus}
As for the current content of the corpus, we believe that extending the works included in the corpus could help understand how poetry is recited when considering different styles. In the current status of the corpus, the text is rather traditional in terms of content, but also in terms of the meter. Besides, if we include readings from other readers, we will have a wider spectrum of readers, and that will help reduce possible reader-specific idiosyncrasies in our analyses.


Following the direction of multimodality, it would be very interesting to include visual information, as poetry is recited in many cases. Following this, the visual information could be integrated and this would make the corpus even more useful than what it currently is.

\bibliographystyle{apalike}
\bibliography{references}

\end{document}